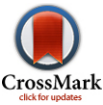

# Perspective Texture Synthesis Based on Improved Energy Optimization

Syed Muhammad Arsalan Bashir[1,2]*, Farhan Ali Khan Ghouri[3]

1 Department of Electrical Engineering, Institute of Space Technology, Islamabad, Pakistan, 2 Quality Management Directorate General, Pakistan Space and Upper Atmosphere Research Commission (SUPARCO), Karachi, Pakistan, 3 Department of Computer Science, University of Karachi, Karachi, Pakistan

**Abstract**

Perspective texture synthesis has great significance in many fields like video editing, scene capturing etc., due to its ability to read and control global feature information. In this paper, we present a novel example-based, specifically energy optimization-based algorithm, to synthesize perspective textures. Energy optimization technique is a pixel-based approach, so it's time-consuming. We improve it from two aspects with the purpose of achieving faster synthesis and high quality. Firstly, we change this pixel-based technique by replacing the pixel computation with a little patch. Secondly, we present a novel technique to accelerate searching nearest neighborhoods in energy optimization. Using k- means clustering technique to build a search tree to accelerate the search. Hence, we make use of principal component analysis (PCA) technique to reduce dimensions of input vectors. The high quality results prove that our approach is feasible. Besides, our proposed algorithm needs shorter time relative to other similar methods.

**Funding:** The authors have no support or funding to report.

**Competing Interests:** The authors have declared that no competing interests exist.

* Email: smab1176@yahoo.com

## Introduction

Texture synthesis has been a basic problem and of wide importance in computer graphics, image processing, and applications of movie or video games. Texture synthesis method is put forward in order to solve the existing out of shape questions in texture mapping. Its aim is explained in [1] as texture synthesis procedure commences with a sample image and tries to produce a texture which is of the same visual appearance as that of the sample.

Perspective textures are textures with global features, such as geometry of the underlying surface, luminance, perspective viewpoint, object distributions, etc. These features exist due to environmental changes (see Figure 1). In this work we propose a novel algorithm for producing perspective texture from an exemplar.

The visual manifestation of an exemplar can be evaluated by the local stability of the texels (TEXture ELment, TEXEL), and global perspective features. Traditional example-based synthesis approaches as in [2–3] synthesize the texture either one pixel or one patch at a point in time, but having the same goal of creating visually similar images with the given example, by comparing the local neighborhoods to maintain coherence of the grown region with nearby pixels. Hence, they could keep the local continuities of the texels well. However, these techniques couldn't acquire good results for perspective textures.

Due to its ability to read and control global feature information of the input texture examples, perspective texture synthesis is becoming more and more important in many fields. Compared with traditional techniques, our work has some significant contributions. We extend an energy optimization technique, i.e. the 2D texture optimization technique, to perspective texture synthesis. Energy optimization is a pixel- based technique, so it consumes lots of time [4–5]. Main motivation of this research is to achieve perspective texture synthesis with global features along with low synthesis time.

To fulfill faster synthesis, and at the same time with high quality, we improve the energy optimization algorithm from the following two aspects:

1) We change it by replacing the pixel-by-pixel computation with a small patch one (2*2 neighborhoods in our implementation).

2) We also present a novel technique to accelerate searching the nearest neighborhoods in energy optimization, a bit similar to the search method in [5]. We use k- means clustering technique to build a search tree first, which is in the same way as in [4]. It allows a good acceleration of the search. But on the basis of this, we make use of PCA technique to reduce the dimensions of input vectors. As the dimensionality of input vector is usually reduced sharply, so is the time to process. Through these improvements, the computation efficiency of energy optimization part is highly increased.

## Related Work

Texture synthesis unravels issues similar to an edge, joint and distortion brought by texture mapping. Now it has essential use in image processing and computer vision. Many researchers have done plenty of work in texture synthesis. In recent years, example-based texture synthesis methods [2,6–9] have achieved great progress. There are two main kinds of example-based synthesis methods: pixel and patch-based [3,10]. The pixel-based synthesis method generates a texture pixel by pixel, often realized by comparing the neighborhood from the output texture to the input while; the patch-based method directly duplicates one patch from the source into the output. Example-based algorithms are also called the neighborhood-based algorithms, because they need to compare and search neighborhoods most similar to the perspective region first.

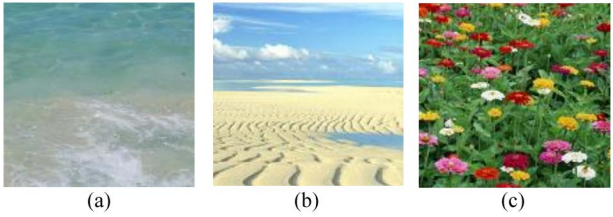

**Figure 1. Some examples of perspective texture, which apparently have visual properties.** These three texture images will be texture samples of our approach, as input, to show the synthesis effect.
doi:10.1371/journal.pone.0110622.g001

All the neighborhood-based techniques consist of two phases [11]: In phase 1, search a most similar neighborhood in the sample with the context region; in phase 2, merge a pixel or patch of the output texture. The search problem always affects the runtime dominantly, so neighborhood search for the primary stage is very time-consuming, hence it's necessary to speed up the search. In [4] authors chose to build a k-means tree to accelerate this process, and in [5] uses Approximate Nearest Neighborhood (ANN) to resolve the nearest neighborhood search problem, and it is effective. In our approach, we used the PCA technique to scale back the dimensions of input on the basis of k- means tree to progressively speed up the nearest neighborhood search. For the second stage, patch merging is optimized using dynamic graph cuts and programming. Although these techniques have achieved significant progress, and many attempts have been made to alleviate the appearance of broken features near the boundary of two neighboring patches, it is still a vital problem.

Latest technique for texture synthesis and image recognition include use of Hidden Markov Model (HMM) in visual vocabulary based computer vision tasks as in [12]. In [12], the only constraint is the maintenance of large indexing structures using a single server subject, hence an optimization is proposed in [13] where they modified the duplicate visual search to index and process millions of images over multiple servers. Using vocabulary based system where the system compares the sample with the vocabulary images; the sparse data error can be avoided.

In the later part of this paper, we introduce some theoretical knowledge about the definition of the slant and tilt angle, a texel scale evaluation technique of example textures, energy optimization, histogram matching techniques and the details of the proposed algorithm in Section 3. Section 4 shows the results of some examples in experiments using our approach. We finally conclude the paper and carry out the course of future work in Section 5.

## Methodology

### 1 Scale Evaluation

For avoiding some situations we can't imagine, we presume all the texture images studied in this paper are locally planar, so that the projective geometry is the only reason to produce texture variations. In our approach, to calculate the relative scale divergence among the texels, we try to estimate visual properties of the texture, and eventually obtain the scale map so that we can perform energy optimization algorithm to achieve texture synthesis.

The scale plot of a texture image has a direct relationship with its surface orientation. We introduce surface orientation using an observer centered spherical coordinate system as in [14] and [15], it has two parameters of tilt ($\tau$) and slant ($\sigma$), and consider any possible optical projections of a round disk with different directions with respect to the sight line as in [14] (See Figure 2). The slant of the circle in 3D space is used to find the aspect ratio of the projected eclipse, whereas orientation of the ellipse within the image plane is given by the tilt. Figure 3(a) shows the definition of tilt and slant angle in 3D space [14,16].

The origin of the image plane is set to the bottom-left corner of the input image, as shown in Figure 3(b). Moreover, through mirroring or flipping operations, largest scale texel is transformed to the origin, whereas smallest one is transformed at the top-right corner. Also, it is presumed that only the slant angle $\sigma$ determines the highest and lowest scale values. So after the user provides the slant and tilt angles, we can begin calculating the scale values by

$$S_{max} = 1/\cos\sigma, \; S_{min} = \cos\sigma, \; S_\Delta = S_{max} - S_{min} \quad (1)$$

Where, the peak scale value is given by $S_{max}$ and smallest value is given by $S_{min}$, and the difference between the two is given by $S_\Delta$. Then,

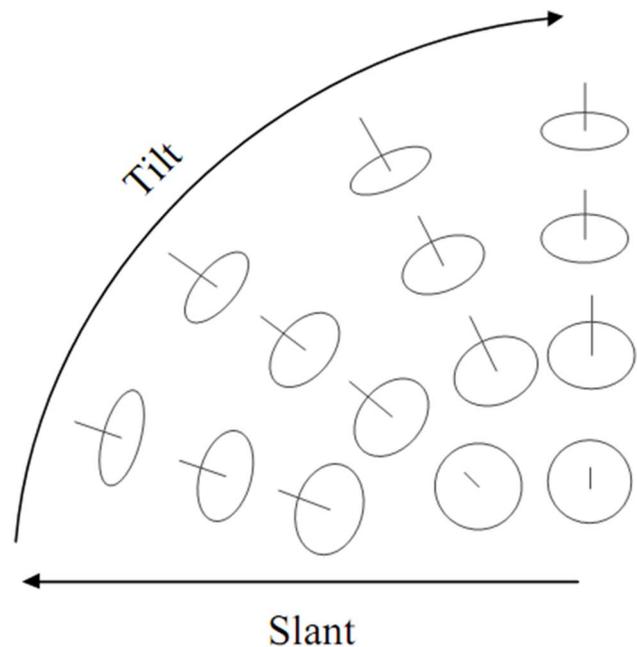

**Figure 2. Texture scale variation using viewer centered spherical coordinate system.** Tilt components of surface orientation is illustrated using a set of round patches arranged on a sphere. Central line at each patch shows the direction of the surface normal.
doi:10.1371/journal.pone.0110622.g002

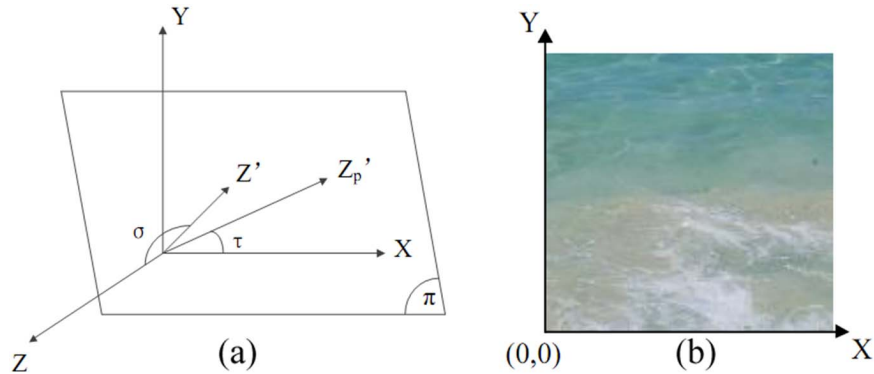

**Figure 3. Definition of slant and tilt angle from 3D space and image coordination.** (a) XY is image plane; Z is normal of image plane; π is texture plane; Z′ is normal of texture plane; Zp′ is the projection of Z′ on XY; σ is the slant angle (between Z and Z′); τ is the tilt angle (between X and Zp′). (b) Image coordination of a texture.
doi:10.1371/journal.pone.0110622.g003

$$Y_{max} = \left(w - 1 - \frac{w}{2}\right)\sin\tau + \left(h - 1 - \frac{h}{2}\right)\cos\tau$$
$$Y_{min} = -Y_{max}$$
$$Y_\Delta = Y_{max} - Y_{min}$$
(2)

Where, h and w respectively indicates the height and width of the input texture image, $Y_{max}$ and $Y_{min}$ are respectively the maximum and minimum Y-coordinate value of the image plane points on the texture plane. $Y_\Delta$ is the difference between the two. Then each point on the image plane has a local scale $S(p)$ given by:

$$p(Y_{proj}) = \left(p(x) - \frac{w}{2}\right)\sin\tau + \left(p(y) - \frac{h}{2}\right)\cos\tau$$
$$S(p) = Y_{max} - \frac{(p(Y_{proj}) - Y_{min})S_\Delta}{Y_\Delta}$$
(3)

Where, $p(x)$ and $p(y)$ represent the coordinates of point p on the image, $p(Y_{proj})$ indicates the texture projected, the image point Y-coordinate values. We can obtain smooth scale maps with this method (for example, in Figure 4).

## 2 Energy Optimization

**2.1 Energy Optimization Method.** Energy of a single synthesized neighborhood is given by the distance to a proximate neighborhood in an input, additionally called texture energy [4]. The sum of individual neighborhood energies then gives the total energy of the synthesized texture. Comparing local neighborhoods in two textures gives the synthesized texture energy in reference to an input texture. This is based on Markov Random Field (MRF) and is based on similarity criterion, which is used in various local pixel-based synthesis techniques. A global metric can be formed using all the locally defined similarity measures, which is used to optimize the whole texture. An iterative algorithm similar to Expectation Maximization (EM) is used to optimize the texture energy [17]. But here, in this paper, we sort it into two phases i.e. optimization phase and search phase. Optimization phase is mainly to compute new output, and search phase is to search the nearest neighborhoods from input, energy optimization performs by iteratively proceeding these two phases.

Denoting the output texture by X (for which we evaluate texture energy) and the input sample texture by Z. Then texture energy $E_T$ over X is given by

$$E_T(x; \{z_p\}) = \sum_p \|x_p - z_p\|^2 \qquad (4)$$

Where, x is the vectorized version of X. For a pre-assigned neighborhood width of w, $N_p$ represents the neighborhood centered on any pixel p, and $x_p$ denotes the sub-vector which represents the pixels in $N_p$. $z_p$ gives the vectorized pixel neighborhood in Z whose appearance is most similar to $x_p$ under the Euclidean norm. So the target of optimization is to minimize energy, i.e. minimize (4).

Using a course resolution input texture is synthesized and then up-sampled to a higher resolution using interpolation; this serves as initialization of texture. An initial estimate of the texture is iteratively refined by decreasing texture energy in iterations, to perform the texture synthesis using (4). With texture initialization given as X, nearest input neighborhood $z_p$ is determined in accordance to each neighborhood output $x_p$. Then X is the texture that minimizes the energy in (4); here x is treated as real-valued continuous vector variable. After this update step x is changed, hence the set of nearest neighboring input $\{z_p\}$ may change also because x is changed. So the process is iteratively repeated until we reach convergence i.e. $\{z_p\}$ becomes constant.

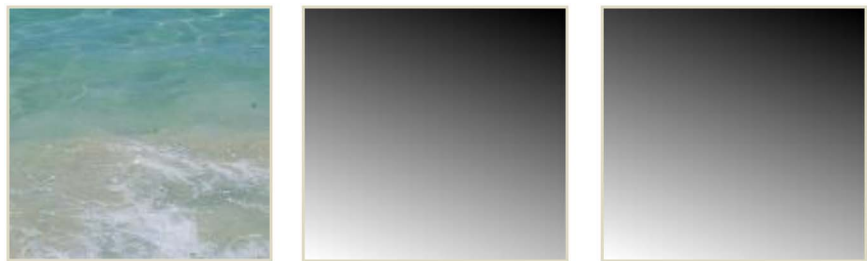

**Figure 4. We obtain scale maps for the given tilt and slant.** From the left to the right: the input example, scale map with σ = 30° and τ = 18°, scale map with σ = 60° and τ = 60°.
doi:10.1371/journal.pone.0110622.g004

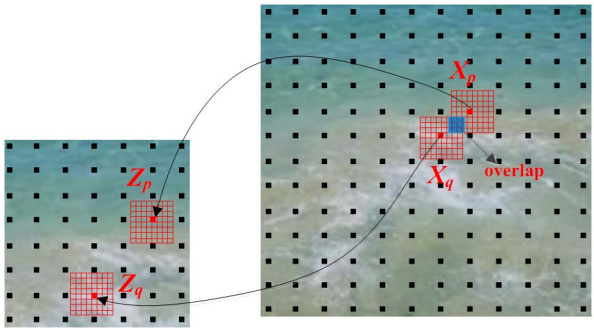

**Figure 5. Energy optimization process on sparse.**
doi:10.1371/journal.pone.0110622.g005

To obtain a more ideal result, we use an energy signal which is robust by replacing the term $\|x_p - z_p\|^2$ in (4) with $\|x_p - z_p\|^r$, where r<2. When we choose r = 0.8, it has the best effect, and is more robust against outliers. So we change function (4) to be

$$E_T(x; \{z_p\}) = \sum_p \|x_p - z_p\|^r \quad (5)$$

To minimize the energy we use Iteratively Re-weighted Least Squares (IRLS), so we have

$$E_T(x; \{z_p\}) = \sum_p \|x_p - z_p\|^r = \sum_p \|x_p - z_p\|^{r-2} \cdot \|x_p - z_p\|^2 \quad (6)$$

$$\text{If } W_p = \|x_p - z_p\|^{r-2} \quad (7)$$

$$\text{Then } E_T(x; \{z_p\}) = \sum_p W_p \cdot \|x_p - z_p\|^2 \quad (8)$$

Energy optimization with respect to function (4) is transferred to minimizing (8).

**2.2 Energy Optimization Improvement.** Through doing a number of experiments and deeply analyzing energy optimization method. We find that there exists a big problem that it is very time-consuming. According to the weakness of this method, we try to improve it from two aspects.

Firstly, in the search phase, searching a nearest neighborhood affects the efficiency of the optimization. In [4], a building *k*-means clustering tree approach is performed. Hierarchical clustering is used to form a tree structure, to organize the input neighborhoods [8,18–19]. Starting with the root node, k-mean clustering is performed with k = 4, for all the neighboring inputs contained in that node. Then k sub-nodes are created corresponding to k clusters and the tree for each sub-node is built recursively. When the neighborhood numbers in a node falls below a predefined threshold (1% of the total), recursion stops. In our design, we are using a threshold less than k (recursion stops when neighboring numbers in a node falls below k). We still choose to use this approach for searching the nearest neighborhood. But due to that the operation is in high-dimensional space, we apply PCA technique to reduce dimensions before the input is processed. This effectively accelerates the search process. In our implementation, to maintain 95% variance, we kept a sufficient number of coefficients in the design. As for textures with 8*8 neighborhoods in RGBS (R, G, B, and Scale channels), the dimensionality is decreased rapidly from 256 to about 20–45. So through this, the synthesis process is obviously accelerated.

Secondly, although this algorithm is based on global optimization, it is based on pixel-by-pixel computation, similar to a pixel-based method, so it is very time-consuming. In [4], it is processed by choosing points on the sparse with intervals of w/4, where w is the fixed neighborhood width; process on the sparse can be seen in Figure 5. It improves a little efficiency, but still needs much time. On the basis of this, we replace every pixel process with a small patch, without changing the computation method of every point. In this way, the number of neighborhoods which wholly contain all the points in a patch is much fewer relative to a single pixel, so the computation is much faster. We deal with all the points in a patch at one time and then move to the next one. In this thought, the only parameter controlled by the user is the size of the block and it concerns the quality and efficiency of method. Through repeated experiments and analysis, we found that 2*2 square pixels block has the best effectiveness. By this time, the result is of the best quality and better than the single-pixel process. So in our implementation, we choose the patch size to be 2*2 square pixels.

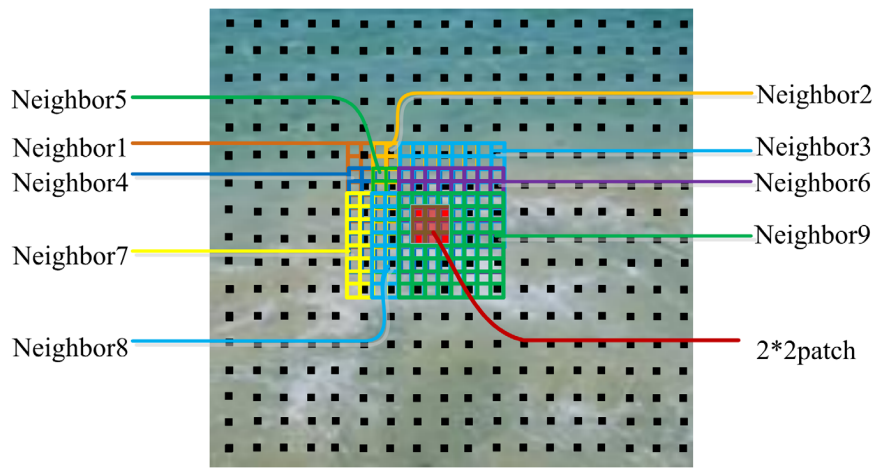

**Figure 6. Neighborhood number: An example of replacing pixel-based computation with 2*2 patch.**
doi:10.1371/journal.pone.0110622.g006

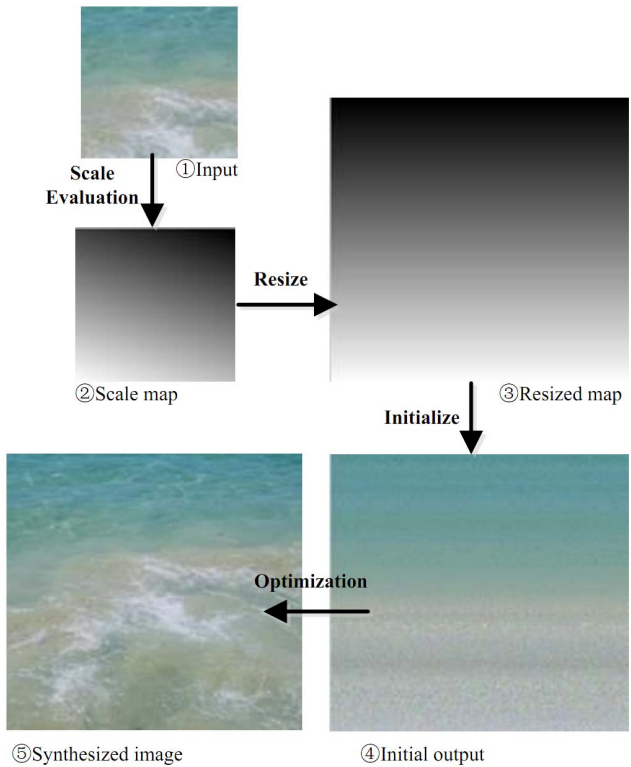

Figure 7. The flow-process diagram of our algorithm.
doi:10.1371/journal.pone.0110622.g007

For example, as shown in Figure 6, for textures with 8*8 square pixels neighborhood, every point in a patch has 4*4 = 16 neighborhoods which contain it, so for four points in a patch there are 16*4 = 64 ones. However, when we process it one patch at a time, for four points in a patch there are just 9 neighborhoods needed. So we reduce the number of neighborhoods which entirely contain the patch relative to a single point, and in like manner time can be saved.

After the improvement, time needed for this algorithm is sharply reduced. Simultaneously, there is also an enhancement for the quality of the result, which will be shown and compared in detail in Section 4.

## 3 Histogram Matching

Sometimes optimization process leads to a wrong convergence, as energy function measures the only similarity of the local neighborhoods, and it does not account for any global statistics. We try to solve this problem by changing the weights in the equation (8) as:

$$W_p = \frac{w_p}{1 + \sum_{j=1}^{k} \max\left[0, H_{x,j}(b_j(z_p)) - H_{z,j}(b_j(z_p))\right]} \quad (9)$$

Where, the number of bins in histogram is denoted by k, jth histogram of the synthesized result and the example is given by $H_{x,j}$ and $H_{z,j}$, respectively, and the value of bin in a histogram H is given by H(b). For a color c, $b_j(c)$ gives the bin containing c in the histograms $H_{x,j}$ and $H_{z,j}$. Through this step, it is ensured that the sample is matched by certain histograms of the output texture. We used Histogram matching technique to reduce the data sparse problem which is caused when patches are used instead of pixels. In our approach, we used scale map to optimize the energy and reduce redundancy; while the histogram matching is used to preserve the global features.

## 4 Synthesis Procedure

Earlier in this paper, we have introduced the main idea and key techniques of our algorithm. Here we are discussing how to synthesize perspective textures using these techniques.

With regard to perspective texture synthesis, the most critical step is to extract the global perspective features. There is a direct relationship between the scale map of a texture image and its surface orientation. So we make use of texel scale evaluation method to obtain scale maps of the input samples. This is the first step of our process, and also the basis to preserve global properties of the given example.

Then we take the scale map as another channel besides R, G, and B three channels of an image. In order to obtain an image

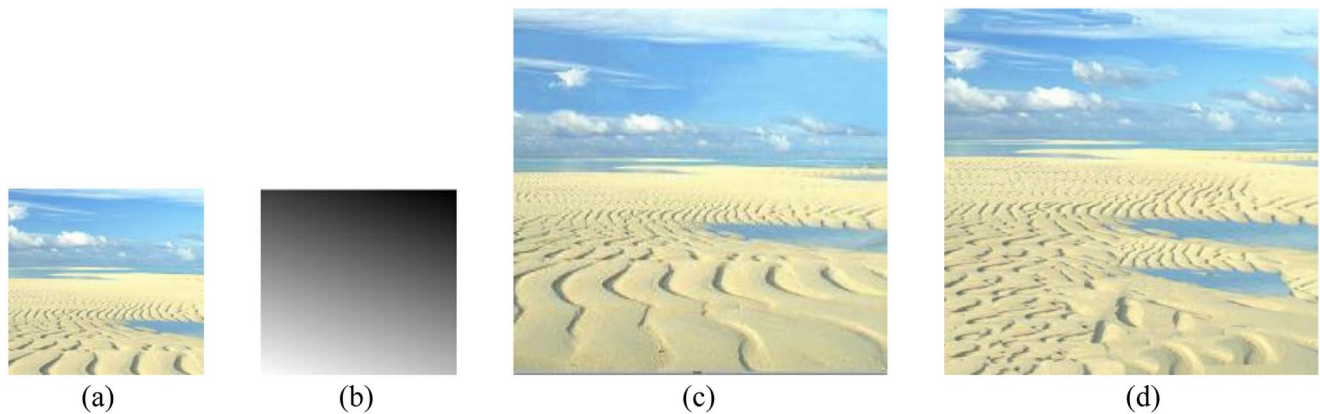

Figure 8. Perspective texture synthesis of a desert image. (a) the input example; (b) scale map with $\sigma = 30°$, $\tau = 18°$; (c) our result; (d) optimization [Kwatra et al. 2005] result.
doi:10.1371/journal.pone.0110622.g008

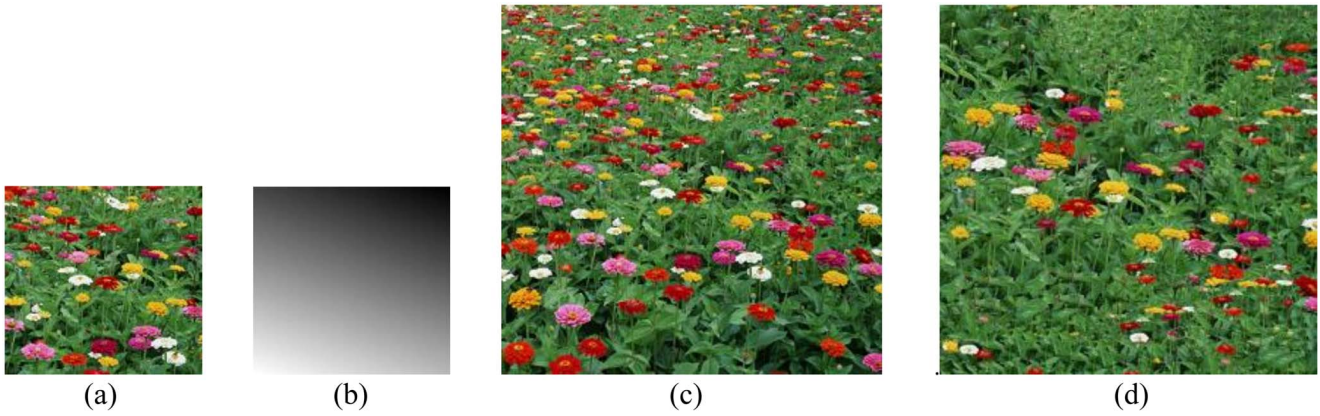

**Figure 9. Perspective texture synthesis of a flower image.** (a) the input example; (b) scale map with $\sigma=60°$, $\tau=20°$; (c) our result; (d) optimization [Kwatra et al. 2005] result.
doi:10.1371/journal.pone.0110622.g009

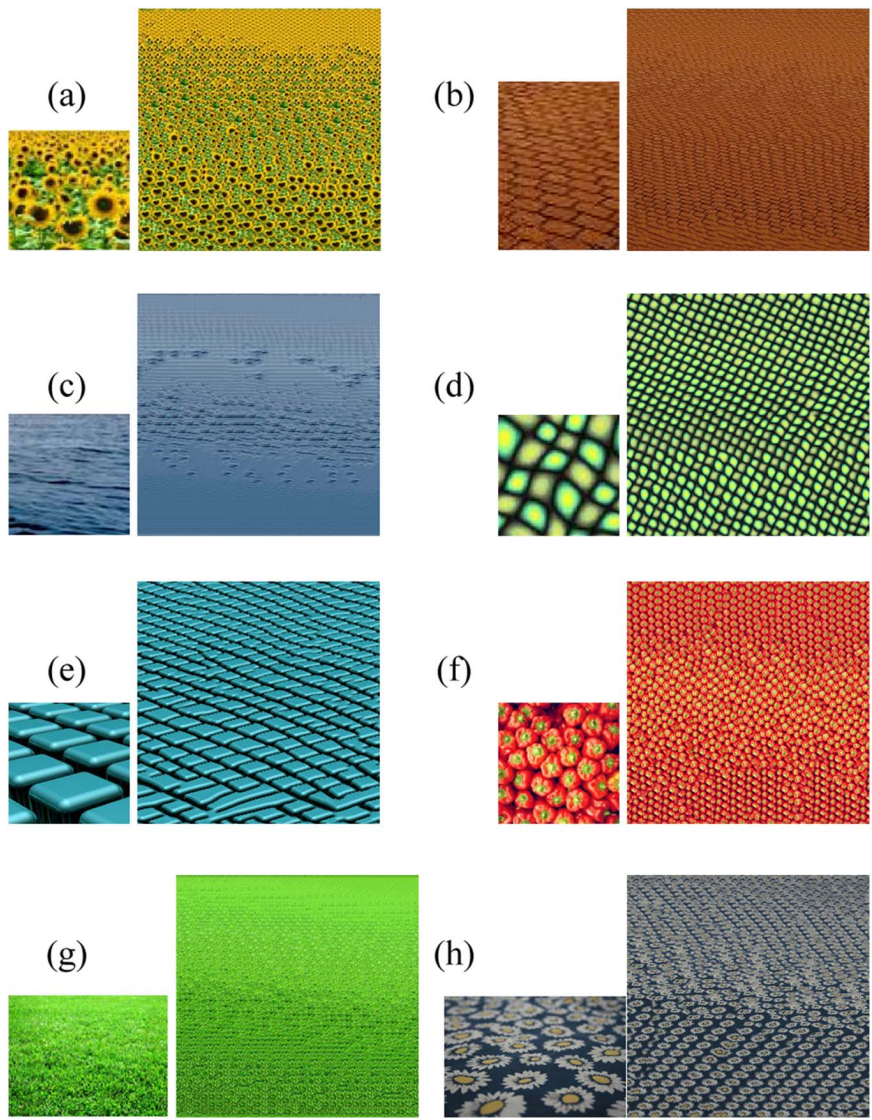

**Figure 10. Perspective texture synthesis of few selected sample textures, for all the textures input is on the left and output is to the right.** Second and third row's last input textures are taken from [Kwatra et al. 2005].
doi:10.1371/journal.pone.0110622.g010

**Table 1.** Efficiency comparison of proposed method with the method in [4].

| Output texture dimension | Our Method | | | Method in [4] | |
|---|---|---|---|---|---|
| | Synthesis Time (min) | Mean (min) | St. Dev. (min) | Mean (min) | St. Dev. (min) |
| 128 pixel×128 pixel (4 different samples) | 0.52 | 0.96 | 0.42 | 2 | 1 |
| | 0.54 | | | | |
| | 0.533 | | | | |
| | 0.57 | | | | |
| | 0.51 | | | | |
| | 0.96 | | | | |
| | 0.92 | | | | |
| | 0.91 | | | | |
| | 0.99 | | | | |
| | 0.95 | | | | |
| | 0.72 | | | | |
| | 0.71 | | | | |
| | 0.76 | | | | |
| | 0.73 | | | | |
| | 0.74 | | | | |
| | 1.62 | | | | |
| | 1.66 | | | | |
| | 1.61 | | | | |
| | 1.62 | | | | |
| | 1.67 | | | | |
| 256 pixel×256 pixel (4 different samples) | 5.74 | 6.55 | 0.75 | 8.5 | 1.5 |
| | 5.67 | | | | |
| | 5.61 | | | | |
| | 5.86 | | | | |
| | 5.65 | | | | |
| | 7.84 | | | | |
| | 7.68 | | | | |
| | 7.73 | | | | |
| | 7.77 | | | | |
| | 7.59 | | | | |
| | 6.33 | | | | |
| | 6.65 | | | | |
| | 6.43 | | | | |
| | 6.51 | | | | |
| | 6.76 | | | | |
| | 6.23 | | | | |
| | 6.21 | | | | |
| | 6.22 | | | | |
| | 6.17 | | | | |
| | 6.29 | | | | |

doi:10.1371/journal.pone.0110622.t001

having global features as the input sample, we resample the scale map in accordance with output size. Re-sampling is performed by comparing the scale values of the input sample and output image that is, choosing the points in the output image having the same scale vales as that of the input sample. Hence the output image points are filled and we get the initial output image and thus direct input image of energy optimization algorithm is achieved.

When getting the initial output image value, we consider the optimization process here. We use the improved energy optimization method to optimize the initial output, and then refined result is obtained. To preserve the global properties, histogram matching is applied to enhance the global features of the synthesized texture to match the sample.

To sum up, our novel sample and energy optimization- based texture synthesis algorithm fulfills the synthesis process through four procedures as following:

1. Perspective scale map extraction from the texel in example, to obtain scale map, with the same size as the sample;
2. Resample the feature maps (input) in accordance with the output size required;
3. Compare the values of input sample and output image, choose point values for which scale values are the same in input sample, to fill the output image points, so that the initial output image is achieved;
4. Optimize the initial output image using improved energy optimization algorithm, and then histogram matching technique, finally output the result, i.e. the synthesized perspective texture image. The flow-process diagram of our algorithm is shown in Figure 7.

As we have mentioned before our algorithm is also very effective. We want to distribute the whole process into two parts to analyze the time and efficiency problem. Part 1 is the pre-processing of data, so that initial texture values are produced and input to energy optimization. It is just some simple computation and choice of points. Part 2 is the process of energy optimization. The computation complexity is dominated by part 2, especially search phase. The number of the nearest neighbor calls is linear in nature. Theoretically, the single call time is given by O ($w^2$), where w is the width of the neighborhood. The original calls for single iteration are given by, $O\left(\frac{n_0}{w^2}\right)$ where $n_0$ is the total output texture pixels and w is neighborhood width. After our improvement, calls per iteration are, $O\left(\left(1-\frac{1}{(w')^2}\right)\cdot\frac{n_0}{w^2}\right)$ where w′ is the width of the patch.

## Experiment Results & Discussions

The input example has no size limit and output texture (of course, the input sample cannot be too smaller to tell the texel) for our approach. The three texture images shown in Figure 1 are very representative, which are with rich perspective properties and can meet our experiment need. So we choose them as input samples. Using the improved energy optimization-based algorithm presented above, we finally obtain the synthesis results shown in Figure 8 and Figure 9 (The seawater image result has been proposed before to discuss the process of this algorithm, so is not shown here again). To better comparison, we provide their results for texture optimization algorithm in [4], separately shown in Figure 8(d) and Figure 9(d). The experiments are done on a computer with 2.4 GHz Intel Core i5 processor, 4 GB RAM, and 32- bit Windows 7 system, using MATLAB software to simulate. The given slant and tilt angles are: σ = 30°, τ = 18° and σ = 60°, τ = 20°, respectively.

From the results shown in Figure 8, and Figure 9, our technique successfully preserves the trends of the global texel scale variation,

and the synthesized images are so perfect with respect to examples. It is proved that this algorithm is very feasible, and has a great advantage over the approach in [4]. Figure 10 shows our results for various sample texture synthesis.

As for the efficiency of our algorithm, we can evaluate that it is very effective from the two parts we have distributed above in Section 3, through the experiments. In our experiments, part 1 usually just needs 2∼5 seconds, whereas part 2 consumes much longer time. It's on the level of minute or even ten minutes. So the time we should mainly take into consideration is part 2, i.e. the energy optimization stage. As seen in Table 1 and File S1, the input examples chosen in this article are 105*105, as for this, an average of 6.5±0.75 minutes are needed for multi-level synthesis textures with the size of 256*256 in part 2, whereas for 128*128, synthesis time is 0.96±0.42 minutes. Compared with other methods [4] where 256*256 textures took 8.5±1.5 mins for multi-level synthesis, while for 128*128 textures it was 2±1 mins, hence our method consumes less time for the same specifications.

## Conclusion

Our approach has achieved great success, and it contributes a bit to the perspective texture synthesis field. The method relies on the scale map to guide, energy optimization and histogram matching to preserve global properties of the sample, so that the process of perspective texture synthesis can be realized. Although we need the user to give the slant and tilt angles during the synthesis process, this perspective texture synthesis technique is still very meaningful and useful in many applications. Not only we achieved high quality results, but we also achieved computational efficiency compared to other methods.

In the future work, we will try our best to study on the synthesis without user assistance, that is, automatic technique. Besides, we should pay more attention to applying texel scale evaluation and getting scale maps from locally non-planar texture images. These are all very difficult and challenging.

## Supporting Information

**File S1** **Efficiency analysis using proposed method for 128*128 pixels and 256*256 pixels synthesis.**
(XLSX)

## Acknowledgments

We thank all reviewers and academic editors for their helpful comments and suggestions. We thank Dr. Weiming Dong, Dr. Vivek Kwatra and Dr. Alexei A. Efros for sharing their results and texture samples on their web sites.

## Author Contributions

Conceived and designed the experiments: SMAB FAKG. Performed the experiments: SMAB FAKG. Analyzed the data: SMAB FAKG. Contributed reagents/materials/analysis tools: SMAB FAKG. Contributed to the writing of the manuscript: SMAB FAKG. Selected additional data set for implementation: SMAB. Acquired permission for authors for including their textures in the paper: FAKG.

## References

1. Ashikhmin M (2001) Synthesizing natural textures. Proceedings of the 2001 symposium on Interactive 3D graphics USA ACM: 217–226.
2. Efros AA, Leung TK (1999) Texture Synthesis by Non-parametric Sampling. IEEE International Conference on Computer Vision Greece: 1033–1038.
3. Kwatra V, Schodl A, Essa I, Turk G, Bobick A (2003) Graphcut textures: image and video synthesis using graph cuts. ACM Trans Graph 22(3): 277–286.
4. Kwatra V, Essa I, Bobick A, Kwatra N (2005) Texture optimization for example-based synthesis. ACM Trans Graph 24(3): 795–802.
5. Kopf J, Fu C-W, Cohen-Or D, Deussen O, Lischinski D, et al. (2007) Solid texture synthesis from 2D exemplars. ACM Trans Graph 26(3): 2.

6. Wei L-Y, Levoy M (2000) Fast texture synthesis using tree-structured vector quantization. Proceedings of the 27th annual conference on Computer graphics and interactive techniques USA: 479–488.
7. Liang L, Liu C, Xu Y-Q, Guo B, Shum H-Y (2001) Real-time texture synthesis by patch-based sampling. ACM Trans Graph 20(3): 127–150.
8. Johnson S (1967) Hierarchical clustering schemes. Psychometrika 32(3): 241–254.
9. Zhang J, Zhou K, Velho L, Guo B, Shum H-Y (2003) Synthesis of progressively-variant textures on arbitrary surfaces. ACM Trans Graph 22(3): 295–302.
10. Efros AA, Freeman WT (2001) Image quilting for texture synthesis and transfer. Proceedings of the 28th annual conference on Computer graphics and interactive techniques USA: 341–346.
11. Wu Q, Yu Y (2004) Feature matching and deformation for texture synthesis. ACM Trans Graph 23(3): 364–367.
12. Ji R, Yao H, Sun X, Zhong B, Gao W (2010) Towards semantic embedding in visual vocabulary. In IEEE Conference on Computer Vision and Pattern Recognition (CVPR): 918–925.
13. Ji R, Duan LY, Chen J, Xie L, Yao H, et al. (2013) Learning to distribute vocabulary indexing for scalable visual search. IEEE Transactions on Multimedia 15(1): 153–166.
14. Dong W, Zhou N, Paul J-C (2008) Perspective-aware texture analysis and synthesis. Visual Compt 24: 515–523.
15. Norman JF, Todd JT, Norman HF, Clayton AM, McBride TR (2006) Visual discrimination of local surface structure: Slant, tilt, and curvedness. Vision Research 46(6–7): 1057–1069.
16. Plantier J, Lelandais S, Boutte L (2001) A shape from texture method based on local scales extraction: precision and results. International Conference on Image Processing 2: 261–264.
17. McLachlan GJ, Krishnan T (2007) The EM algorithm and extension. Hoboken NJ: Wiley 382.
18. Elkan C (2003) Using the Triangle Inequality to Accelerate k-Means. In T. Fawcett, & N. Mishra (Ed.), Proceedings of the Twentieth International Conference on Machine Learning USA: 147–153.
19. Dellaert F, Kwatra V, Oh SM (2005) Mixture Trees for Modeling and Fast Conditional Sampling with Applications in Vision and Graphics. Proceedings of the 2005 IEEE Computer Society Conference on Computer Vision and Pattern Recognition USA 1: 619–624.